\documentclass{article}

\usepackage[final]{neurips_2022}
\usepackage[utf8]{inputenc} 
\usepackage[T1]{fontenc}    
\usepackage{hyperref}       
\usepackage{url}            
\usepackage{booktabs}       
\usepackage{amsfonts}       
\usepackage{nicefrac}       
\usepackage{microtype}      
\usepackage{xcolor}         

\usepackage{amsmath,amssymb}
\usepackage{graphicx,wrapfig}
\usepackage{tikz}

\usepackage{algorithm}
\usepackage{algorithmic}
\usepackage{mathtools}

\DeclarePairedDelimiter{\norm}{\lVert}{\rVert} 

\setcitestyle{authoryear,semicolon,open={(},close={)}} 

\title{Reinforcement Learning in System Identification}

\author{
  Jose Antonio Martin H.\\
  \And 
  Oscar Fernández\\
  \And 
  Sergio Pérez\\
  \And 
  Anas Belfadil\\
  \And 
  Cristina Ibanez-Llano\\
  \And 
  Freddy José Perozo Rondón\\
  \And 
  Jose Javier Valle\\
  \And 
  \vspace{5mm}
  Javier Arechalde Pelaz\\
  \vspace{2mm}
  Repsol Technology Lab, Repsol, Spain\\
  \texttt{\{ja.martin.h, oscar.fernandez.v, sergio.perez.m, anas.belfadil,}\\
  \texttt{cristina.ibanez, f.perozo, jjvallea, javier.arechalde\}@repsol.com}
}

\begin{document}

\maketitle

\begin{abstract}
 System identification, also known as learning forward models, transfer functions, system dynamics, etc., has a long tradition both in science and engineering in different fields. Particularly, it is a recurring theme in Reinforcement Learning research, where forward models approximate the state transition function of a Markov Decision Process by learning a mapping function from current state and action to the next state. This problem is commonly defined as a Supervised Learning problem in a direct way. This common approach faces several difficulties due to the inherent complexities of the dynamics to learn, for example, delayed effects, high non-linearity, non-stationarity, partial observability and, more important, error accumulation when using bootstrapped predictions (predictions based on past predictions), over large time horizons. Here we explore the use of Reinforcement Learning in this problem. We elaborate on why and how this problem fits naturally and sound as a Reinforcement Learning problem, and present some experimental results that demonstrate RL is a promising technique to solve these kind of problems.
\end{abstract}

\section{Introduction} \label{Sect:Intro}
Learning forward models has been an active area of research in past decades, with abundant contributions on the application of Machine Learning (ML) techniques to the "system identification problem"~(see for instance, \citealp{Werbos1989, Fu2013, Zhang2014, Abdufattokhov2019, ROEHRL20209195}). 

Particularly, it is a recurring topic of research within Reinforcement Learning~(RL, see~\citealp{sutton1991dyna,sutton1998introduction,polydoros2017survey,DBLP:journals/corr/abs-2006-16712}), where forward models usually represent the transition function $s_{t+1} = \mathcal{T}(s_t,a_t)$ of some Markov Decision Process (MDP). We denote an MDP as a tuple $\mathcal{M} = (\mathcal{S}, \mathcal{A}, \mathcal{T}, \mathcal{R})$, where $\mathcal{S}$ denotes the state space, $\mathcal{A}$ denotes the action space, $\mathcal{T}$ denotes the transition function and $\mathcal{R}$ denotes the reward function. Thus, $s_{t+1}=\mathcal{T}(s_t,a_t)$ represents the immediate state after the evolution of the system, starting at time $t$ with state $s_t$ and conditioned by an action $a_t$. Hence, $\mathcal{T}$ can be represented by a deterministic or stochastic mapping function $s_{t+1}= \mathcal{F}(s_t,a_t)$.

Learning a forward model, i.e., learning a mapping function $\hat{s}_{t+1}= f(s_t,a_t)$, is commonly defined as a Supervised Learning problem in a direct way~\citep{jordan1992forward,DBLP:journals/corr/abs-2006-16712}, since there are well-defined observations $\mathbf{X} = \{(s_t,a_t),...\}$, labels $\mathbf{y} = \{s_{t+1},...\}$, and a loss, e.g.,  $\mathcal{L}=||f(s_t,a_t)-s_{t+1}||_{2}$, which are the core parts of such problems. However, we must note that what we are \textit{given} is a set of observations solely, and it is just a reasonable \textit{assumption} defining the labels as $\mathbf{y} = \{s_{t+1},...\}$. In practice, this approach faces several difficulties due to the inherent complexities of the dynamics to learn, thus we propose that the problem of learning forward models can be more naturally defined, and indeed effectively solved, as an RL problem.

\section{Motivation and Problem definition} \label{sect:Motivation}
\paragraph{Why learning forward models with RL?\label{intro:complexities}} The domains, tasks, and problems to which forward modeling is being applied are of increasing complexity, including time delayed dynamical effects, high degree of non-linearity, partial observability (POMDPs), and in general, complex dynamics. This situation raised the need of additional techniques to adapt the Supervised Learning framework to deal with this increasing complexity, see for instance \citep{oh2015action,silver2017predictron,xiao2019learning,lambert2021learning,lambert2022investigating}:

\begin{enumerate}
    \item Rollout testing for beyond single step learning robustness.
    \item Loss accumulation over rollouts for large horizon prediction.
    \item Recurrent networks, frame-stacking, or neural Turing Machines for partial observability.
    \item Curriculum learning over increasing horizons to aid learning convergence.
    \item Data augmentation to aid learning symmetries in data.
    \item Ensembles of stochastic neural networks to increase the prediction accuracy and reduce bias.
\end{enumerate}
On the other hand, RL has intrinsic features that provides a natural way to deal with many of those complexities and provides even more:
\begin{enumerate}
    \item Rollout learning by working with episodic tasks.
    \item Minimization of (compounding) error accumulation by optimizing in the long-run.
    \item Stochastic scenarios.
    \item Partial observability.
    \item Solving the sequential credit assignment problem.
    \item Continuous learning from new experience without requiring a full retraining.
\end{enumerate}

\paragraph{Unreasonable extra search cost?} Solving a regression problem by RL has an extra cost due to the required exploration, i.e., searching for a point $y_i \in \mathbf{y}$ (label) that is indeed already known. Thus, why solving this problem with RL? Is it an unreasonable cost? May we have now two problems instead of one?
 
Let's suppose that we run bootstrapped rollouts of certain length (time horizon), while minimizing the errors between each predicted point $x_t=f(s_t,a_t)$ and its corresponding "true target" (label) $y_t \in \mathbf{y}$. Since the predictions in the rollouts trajectories are \textit{sequentially dependent} (by bootstrapping), then we face a temporal credit assignment problem, q.e., the devils behind the compounding error. The RL framework deals with this issue naturally through the Bellman's optimality criteria~(\citealt{bellman2003dynamic}). Also, we shall emphasize that we are not given "true targets", instead we are \textit{assuming} that the next observed states are the targets. Moreover, since we are relying on bootstrapping, then, in the Supervised Learning setting, we need the corresponding observation points $x_t$ for the predicted labels $\hat{y_t}=f(x_t)$, however, these are not in the dataset, since $x_t$ are predictions as well.

Finally, this extra search cost allows to build a $Q$-function that provides a prediction on the expected approximation error we will commit by following the optimal policy as well as being able to provide hedged-predictions~\citep{gammerman2007hedging} on the dynamics evolution.

\paragraph{How to learn forward models with RL?} Learning forward models with RL can be achieved directly just by translating a regression problem to an RL problem, where the observation (state) is formed by the current state of the system and the previous observed action, the actions of the agent represent the predictions of the next state, and the reward signal is just the negated total/cumulative prediction error. Note that now we don't have to assume that next observations are targets.

More formally, given an MDP, $\mathcal{M} = (\mathcal{S}, \mathcal{A}, \mathcal{T}, \mathcal{R})$, the forward learning problem is defined as the MDP $\mathcal{M_F} = (\mathcal{S_F}, \mathcal{A_F}, \mathcal{D|O}, \mathcal{L_F})$, 
where $\mathcal{D|O}$ refers to a time series of transitions stored in a dataset ($\mathcal{D}$) or observed ($\mathcal{O}$) from a real world process, 
$\mathcal{L_F}=||s_{t+1} - \hat{s}_{t+1}||$ is a loss function of the true observed next state ($s_{t+1}\leftarrow \mathcal{D|O}$) and the predicted next state ($\hat{s}_{t+1}$), 
$\mathcal{S_F}$ denotes the state space formed by tuples of $(s,a)\in \mathcal{S}\times\mathcal{A}$, 
and $\mathcal{A_F}$ is the action space defined by elements $\Delta s = \hat{s}_{t+1} - \hat{s}_{t}$, where $s\in S$. Figure~\ref{fig:rl:forwardmodel} shows the diagrams for the MDPs $\mathcal{M}$ and $\mathcal{M_F}$, respectively.

\tikzstyle{block} = [rectangle, draw, text width=7em, text centered, rounded corners, minimum height=2em]
\tikzstyle{line} = [draw, -latex]
\begin{figure}[htb]
    \centering
    \begin{tikzpicture}[node distance = 5em, auto, thick]
    \node [block] (Agent) {$agent$\\$a_{t+1}\leftarrow f(s)$};
    \node [block, below of=Agent] (Environment) {$environment$\\$s_{t+1}\leftarrow\mathcal{T}(s_t,a)$};
    
     \path [line] (Agent.0) --++ (2.5em,0em) |- node [near start, left]{$a$} (Environment.0);
     \path [line] (Environment.190) --++ (-4em,0em) |- node [] {$s$} (Agent.170);
     \path [line] (Environment.170) --++ (-2.5em,0em) |- node [near start, right] {$r$} (Agent.190);
    \end{tikzpicture}
    \begin{tikzpicture}[node distance = 5em, auto, thick]
    \node [block] (Agent) {$agent$\\$\Delta s \leftarrow f(\hat{s},a)$};
    \node [block, below of=Agent] (Environment) {$environment$\\$\hat{s}_{t+1}=\hat{s_t}+\Delta s$\\$a_{t+1}\leftarrow \mathcal{D|O}$};
    
     \path [line] (Agent.0) --++ (2.5em,0em) |- node [near start, left]{$\Delta s$} (Environment.0);
     \path [line] (Environment.190) --++ (-4em,0em) |- node [] {$(\hat{s},a)$} (Agent.170);
     \path [line] (Environment.170) --++ (-2.5em,0em) |- node [near start, right] {$\mathcal{L}$} (Agent.190);
    \end{tikzpicture}
   
    \caption{The typical RL setting (left). RL flow for learning a forward model (right).}
    \label{fig:rl:forwardmodel}
\end{figure}
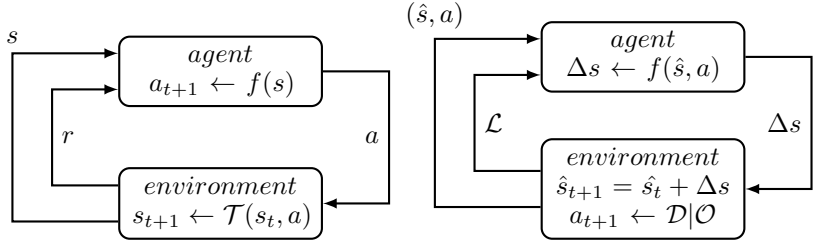

Hence, to apply RL to the forward model learning problem, we just need a "transposition" of the problem and its observations, where the final objective is to obtain, by RL, a policy $\pi_{\hat{\mathcal{M}}}$ on the approximated dynamics $\pi_{\mathcal{M_F}} \leftarrow \mathcal{M_F}$, such that $\pi_{\hat{\mathcal{M}}}$ is as close as possible to the optimal policy $\pi_{\mathcal{M}}$ in the original MDP, as shown in Equations~\ref{eq:fwd:policy1} to \ref{eq:fwd:policy4}.
\begin{align}
    \pi_{\mathcal{M}} \leftarrow \mathcal{M} &= (\mathcal{S}, \mathcal{A}, \mathcal{T}, \mathcal{R}) \label{eq:fwd:policy1} \\
    \pi_{\mathcal{M_F}} \leftarrow \mathcal{M_F} &= (\mathcal{S_F}, \mathcal{A_F}, \mathcal{D|O}, \mathcal{L_F}) \label{eq:fwd:policy2} \\
    \pi_{\hat{\mathcal{M}}} \leftarrow \hat{\mathcal{M}} &= (\mathcal{S}, \mathcal{A}, \pi_{\mathcal{M_F}}, \mathcal{R}) \label{eq:fwd:policy3} \\
    ||\pi_{\mathcal{M}}-\pi_{\hat{\mathcal{M}}}|| \: &\propto \: ||\mathcal{T}-\pi_{\mathcal{M_F}}|| \label{eq:fwd:policy4}
\end{align}

\section{Derivation and method} \label{sect:Method} 
	Starting from the one-step definition of a forward model (Eq.~\ref{eq:fwd:onestep}):
	\begin{equation}
		\label{eq:fwd:onestep}
		\hat{s}_{t+1}= f(s_t,a_t),
	\end{equation}
	where $s_t$ is the \textit{state} of the system at time step $t$ and $a_t$ being the control actions applied to the system at $t$, we obtain a one-step (\textit{forward}) prediction of the next \textit{state} $\hat{s}_{t+1}$ of the system. Learning such a forward model $f(s,a)$ implies learning the approximator function $f\approx\mathcal{T}$ with any statistical or ML method (e.g., by fitting the parameters $\theta$ of a parametric estimator $f_{\theta}$). For this, a loss function is defined as $\mathcal{L}$ (Eq.~\ref{eq:loss}), 
	\begin{equation}
		\label{eq:loss}
		\mathcal{L} = ||s_{t+1} - \hat{s}_{t+1}||,
	\end{equation}
	where $\mathcal{L}$ is minimized during training, for instance, using gradient descent techniques by computing the gradient $\nabla_{\theta}\mathcal{L}$. Thus, the problem setting naturally appears as a Supervised Learning problem (regression) $\hat{y}_i=f(x_i)$, where the collection of inputs $\mathbf{X}$ is formed by tuples $x_i= (s_t,a_t)$ and the targets collection (labels) $\mathbf{y}$ are commonly defined by the next observed states $y_i=s_{t+1}$.
   
    With this definition, a Supervised Learning task is completely defined and can be then easily achieved through a "vanilla" ML pipeline. However, as we noted in Section~\ref{intro:complexities}, the results can suffer from several issues, requiring thus, a battery of adhoc methods and other "over complexifications". 
   
   For instance,~\citet{fomoro} gives a short list of common tips to consider. In this work, we consider and apply the following ones: 1.\ predict the deltas from the current state, not the next state, directly, 2.\ use frame-stacking-like window of past observations, 3.\ use data-augmentation techniques, 4.\ use special regularization techniques such as noise or Dropout, 5.\ evaluate the model accuracy using rollout evaluations and 6.\ more importantly, use bootstrapped rollout training (predictions over past predictions).

   From the immediate above list as number 1, by predicting deltas, Eq.~\ref{eq:fwd:onestep} becomes of the following form  (c.f. Eq.~\ref{eq:fwd:onestep:delta}):
   	\begin{align}
		\label{eq:fwd:onestep:delta}
		\hat{s}_{t+1} &= s_t + f(s_t,a_t), \text{ and hence,} \\
	\frac{\Delta \hat{s}}{dt} &= f(s,a),
	\end{align}
   which expresses a well-known recursive relation in the field of Dynamical Systems, by expressing the evolution of the system in terms of its past state plus its derivative (residual).
   
   The last point on the above list, appears as the most difficult part from the learning perspective, however, we claim it is a requirement for a sound and robust definition of a forward model learning problem. Additional features are also inherent to a rollout training when applying RL, for instance, it creates, trough exploration, an implicit data-augmentation over the source data ($\mathcal{D|O}$), which is a common practice done as an extra step in the supervised ML setting, to improve generalization and reduce over-fitting~\citep{Dyk2001,hernandez2018data,Iwana2021}. This implicit data augmentation, instead of enlarging a training dataset from existing data using various translations, acts as a kind of "spatio-temporal" transformation, as in the case of autoencoders, generating new training data "on-the-fly"~\citep{tu2018spatial}, contributing to obtain robust policies to noisy inputs, aid in learning problem symmetries and generalization, and thus helping to avoid overfitting to the fixed dataset or observations.

   \paragraph{Function approximation: Stochastic policies} Focusing on the most recent works on model-based RL~(\citealt{10.5555/3454287.3455409,NEURIPS2020_a322852c}), stochastic Gaussian networks can be used as the function approximator to learn the policy $f\approx\mathcal{T}$, such that:
   \begin{align}
       f &= \frac{\Delta \hat{s}}{dt} \sim \mathcal{N}\left(\mu_{s,a},\,\sigma_{s,a}^{2}\right)
   \end{align}
   This allows to approximate the dynamics through a stochastic model. There is, perhaps, good margin for improvements over Gaussian networks as pointed out by \citet[i.e., beta distribution]{pmlr-v70-chou17a}, however, the original SAC algorithm, which we use for our experiments, uses a squashed Gaussian Normal Policy for the policy network~(see, \citealt{pmlr-v80-haarnoja18b}).
   
   \paragraph{Rollout loss and evaluation} An effective forward model should predict not only the next state, based on the true past state and true action accurately, but should allow as well to simulate the system by running rollouts over its own predictions. The rollout process then implies a bootstrapping process, that is, predicting the next state based on a previous predicted state. Thus, by running a rollout $\mathrm{E}(s_t,(\mathrm{a})_t^{t+h},h)$ of length $h$ simulation steps over a fixed sequence of actions $(\mathrm{a})_t^{t+h}=(a_t...a_{t+h})$ from an initial state $s_t$, we obtain:
	\begin{align}
		\label{eq:rollout}
		\hat{s}_{t+h} &= s_t + \sum_{i=0}^{h} f\left(\hat{s}_{t+i},a_{t+i}\right) \; \biggr\rvert_{\hat{s}_{0}=s_t },
	\end{align}
	obtaining a bootstrapped prediction $\hat{s}_{t+h}$ of length $h$ and a rollout trajectory $\hat{y}= (s_t,\hat{s}_{t+1}, ..., \hat{s}_{t+h})$. Also, equations~\ref{eq:fwd:onestep:delta} to \ref{eq:rollout} define well known recurrent relations that can be seen as the equivalent of recurrent connections, computed with a loop over a sequence input batch where loss is calculated at the end, as well as dynamical system modeling through Neural ODEs (ODENet, \citealt{chen2018neural}), its augmented version~(\citealt{teh2019augmented}), and ResNets~(\citealt{he2016deep}).

	Thus, the rollout loss $\mathcal{L}_E(s_t,(\mathrm{a})_t^{t+h},y,\hat{y},h)$ can then be defined as follows:
	\begin{align}
		\label{eq:rloss}
		\mathcal{L}_E(s_t,(\mathrm{a})_t^{t+h},y,\hat{y},h) &= \sum_{i=1}^h  ||s_{t+i} - \hat{s}_{t+i}||, \; \, s_{t+i} \in y, \hat{s}_{t+i} \in \hat{y}, \nonumber \\
		&= \mathcal{Z}_E(y,\hat{y}),
	\end{align}
	being $\mathcal{Z}_E(y,\hat{y})$ a signal (trajectory) similarity function. Thus, in rollout learning, the network is trained with the rollout loss~(Eq.~\ref{eq:rloss}), instead of the common supervised loss~(c.f. Eq.~\ref{eq:loss}).
	
	The signal similarity function $\mathcal{Z}_E(y,\hat{y})$ is an open choice, however, it changes the problem to be solved, as it is the goal of the RL problem. In particular, here the problem is sequential and thus the objective is to evaluate how "close" (in terms of shape/pattern recognition) is the predicted trajectory vs. the observed trajectory ($\hat{y}$ vs. $y$). We must note that (r)mse for this problem is just a weak approximation, and that doing signal/pattern recognition using (r)mse alone to measure signal/shape similarity is definitely not a good general measure, see for instance~\cite{2019arXiv190205180P}. Also, there are many measures for signal similarity (e.g., KL-divergence, statistical (invariant) moments, signal correlations, etc. ), and usually a combination works better depending on the type of the signals and which signal's features are more important for the problem. 
	
	\paragraph{Policy learning with the Actor-Critic architecture} Finally, a natural way to implement rollout learning is episodic learning, where sequences of transitions are divided in episodes, and rollouts are run over such episodes using loss accumulation for computing the gradient. This approach naturally conducts to thinking on the temporal credit assignment problem~(see, \citealt{minsky1961steps,sutton1984temporal,sutton1998introduction}), which is solved effectively through TD-Learning~(\citealt{sutton1988learning}). That is, to solve a sequential decision problem optimizing a scalar signal over an episodic task (a rollout). For this it looks natural to use the Actor-Critic method~\citep{sutton1998introduction,degris2012off} to train a policy network and, in particular, a stochastic one to predict the deltas (residuals) of the next system state. This can be achieved trough methods like the SAC algorithm~\citep{pmlr-v80-haarnoja18b}.

\section{Experimental evaluation} \label{sect:Expermimental}

We test the proposed approach on three different MuJoCo environments from OpenAI Gym control suite: Hopper-v2, Walker2d-v2, and Halfcheetah-v2. D4RL~\citep{fu2020d4rl} datasets of the MuJoCo environments have been used to train the forward models. A specific Gym environment $\mathcal{M_F}$ has been developed for training over rollouts/episodes of the datasets' time series trajectories. 

All MuJoCo environments used here rely on continuous state and action spaces. Further details on the data preparation can be found in Appendix~\ref{App:datasets}. In MuJoCo, the state variables are divided in two vectors: positions ($qpos$) and velocities ($qvel$). Since positions can be derived from velocities and vice-versa then we opted to predict only the deltas of the positions $\Delta qpos$, and then infer the original velocities of the simulation. The predicted $qpos$ and its corresponding $qvel$ are calculated according to: 
\begin{align}
qpos_{t+1} &= qpos_t + \Delta {qpos} \\
qvel_{t+1} &= \frac{\Delta qpos}{dt}
\end{align}
In order to define the state space $\mathcal{S_F}$, a window (stack) of $w=20$ past time steps observations is used. Every prediction relies on this stack plus the respective taken actions stored in the datasets. At each step, every new prediction is incorporated into $w$ in a FIFO way. Also, for giving more variability to the training data, each episodes starts randomly from a time step in the range $[0,30]$, so that the RL agent is not seeing the same sequences all the time.

We use the Soft Actor-Critic (SAC) algorithm~\citep{pmlr-v80-haarnoja18b} as the base of our trainable agents, thus the actor networks represent the learned models. For all the experiments, the same set of hyperparameters for the Actor and Critic are used: encoders architecture is an MLP of $6$ hidden layers of $512$ elements each with Mish~\citep{misra2019mish} as activation function. Quantile regression~\citep{qregression} is used for the $Q$-function networks (with $n$-\textit{quantiles} $=64$). For each network, and learning step, the batch-size used is $2^{10}$. Networks inputs are min-max scaled based on datasets. Other information about specific training parameters like the number of episodes run per experiment, or gradient steps can be found at Table \ref{table:param-table}. Refer to Appendix \ref{App:hardware} for additional details about hardware and software used.

\begin{table}[tbh]
  \caption{Experiment's training parameters and main test metrics results.}
  \label{table:param-table}
  \centering
  \begin{tabular}{lccrcrrrr}
  
    \toprule
    \multicolumn{1}{l}{Environment}& \multicolumn{4}{c}{Training Parameters}  & \multicolumn{4}{c}{Testing - last values (average)}               \\
    \cmidrule(r){2-5}     \cmidrule(r){6-9}
 &	 Episodes&	 \#Params &	 $dt$ &	Grad. Steps &	Cr. loss&	Act. loss&	rmse\\
 \midrule								        
 Hopper-v2 &	 $12K$ &	 $1425K$ &	 $0.008$ &	$\sim 121K$ &	 $0.49$ &	 $2.05$ &	 $0.09$  \\
 Walker2d-v2 &	 $12K$ &	 $1488K$ &	 $0.008$ &	$\sim 122K$ &	 $5.76$ &	 $10.63$ &	 $0.19$  \\
 Halfcheetah-v2 &	 $12K$ &	 $1488K$ &	$0.05$ &	 $\sim 121K$ &	 $69.06$ &	 $33.00$ &	 $0.28$   \\

    \bottomrule
  \end{tabular}
\end{table}

For these experiments, we used a pseudo-sparse reward signal as follows:
\begin{equation}
\label{eq:reward}
r_t(s,a) =
	\begin{cases}
	 1-\mathcal{Z}_E(y,\hat{y}) (\text{signal similarity function}) & \text{at rollout ends,}  \\
     -\norm{s_t -\hat{s}_t}_{L_2} & \text{otherwise},
    \end{cases}
\end{equation}
being the signal similarity function $\mathcal{Z}_E(y,\hat{y})$:
\begin{equation}\displaystyle
 \mathcal{Z}_E(y,\hat{y}) = 
    \left(1+\norm{y-\hat{y}}_{L_2}\right) 
    \left(1+\norm{\nabla y - \nabla \hat{y}}_{corr}\right)
    \left(1+\norm{y - \hat{y}}_{KL}\right),  
\end{equation}
where, $\norm{y-\hat{y}}_{L_2}$ is the sum of the squared error between true and the predicted values, $\norm{\nabla y - \nabla \hat{y}}_{corr}$ is the correlation distance between the derivatives of both predicted and true values and $\norm{y - \hat{y}}_{KL}$ is the Kullback-Leibler divergence between both true and predicted values, over all the the current rollout steps.

\begin{figure}[tb]
\includegraphics[width=0.32\linewidth,height=80pt]{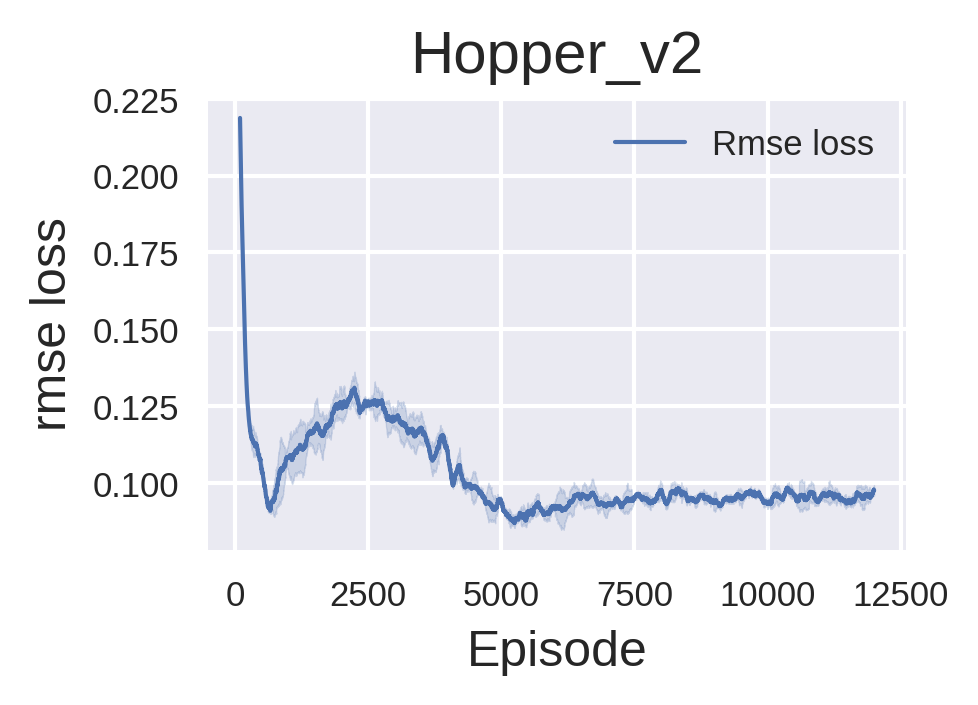}
\includegraphics[width=0.32\linewidth,height=80pt]{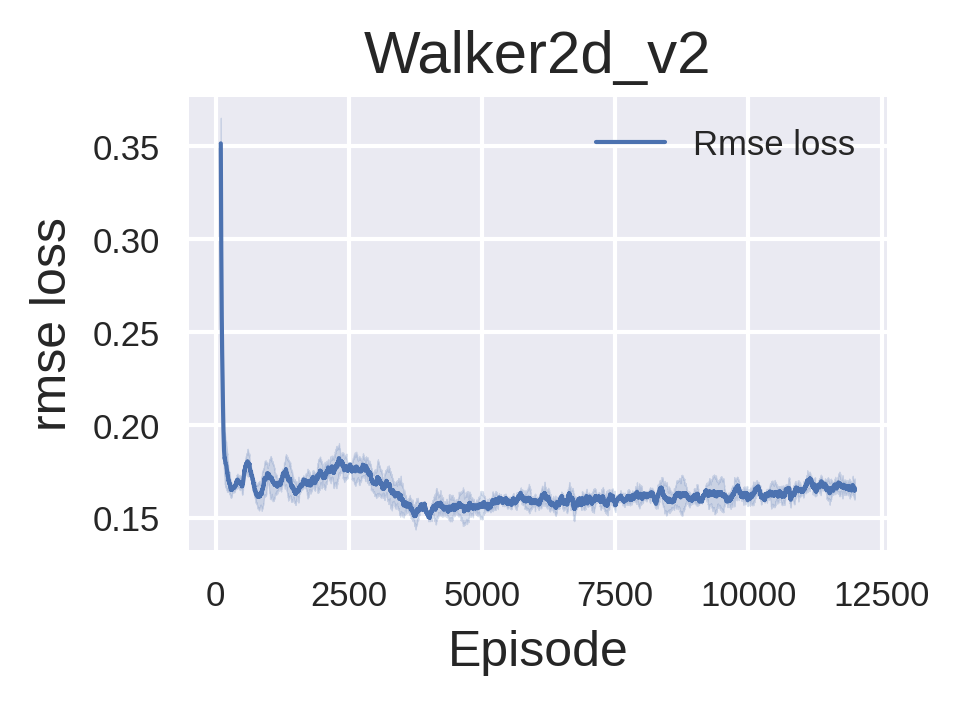}
\includegraphics[width=0.32\linewidth,height=80pt]{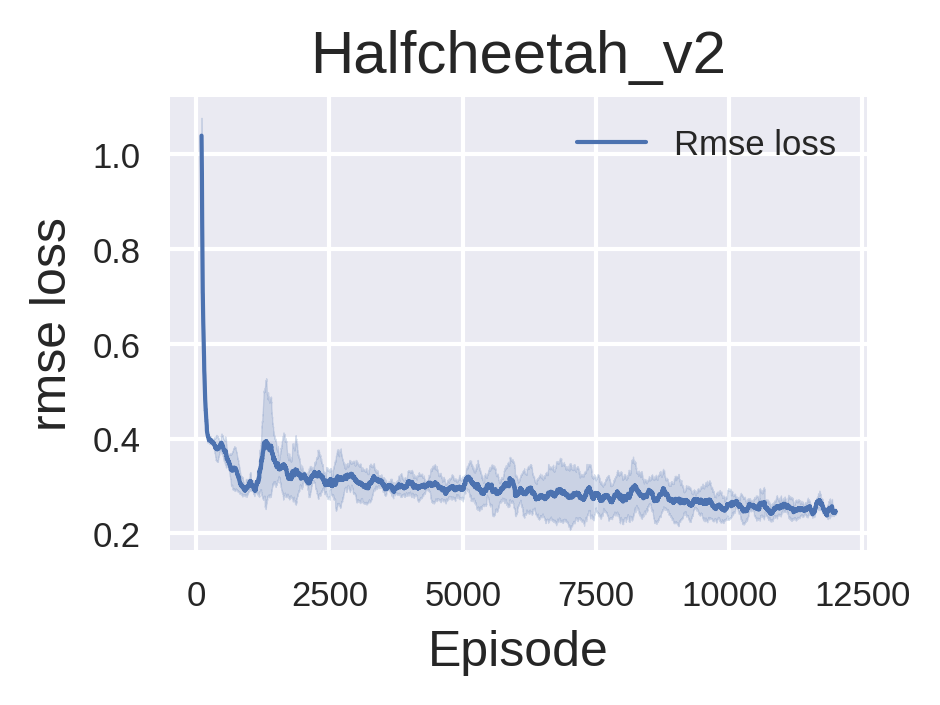}

\includegraphics[width=0.32\linewidth,height=80pt]{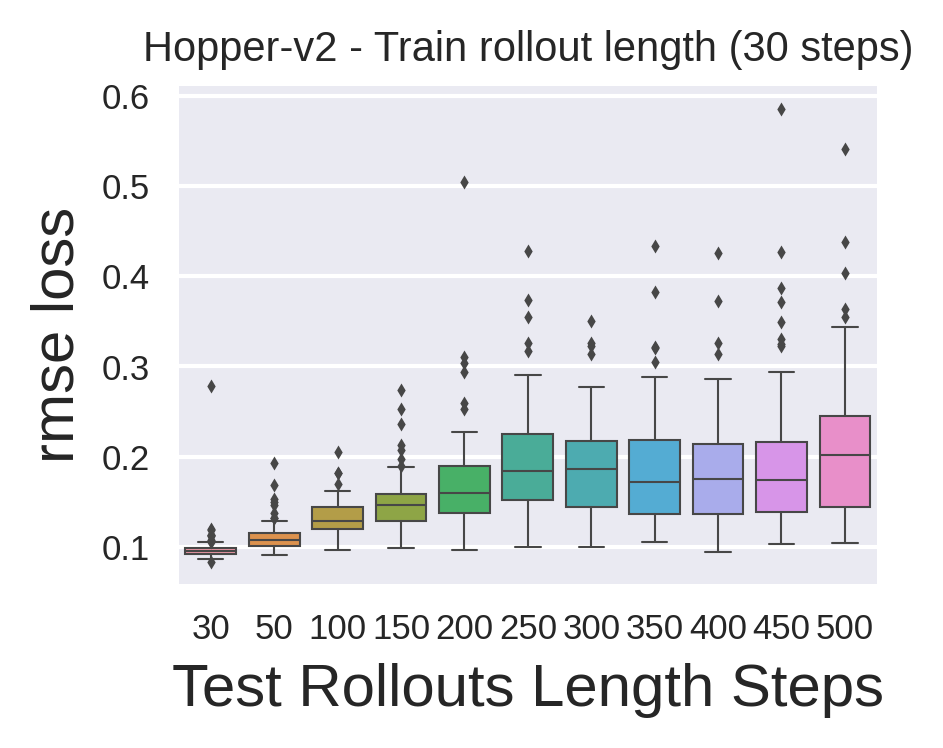}
\includegraphics[width=0.32\linewidth,height=80pt]{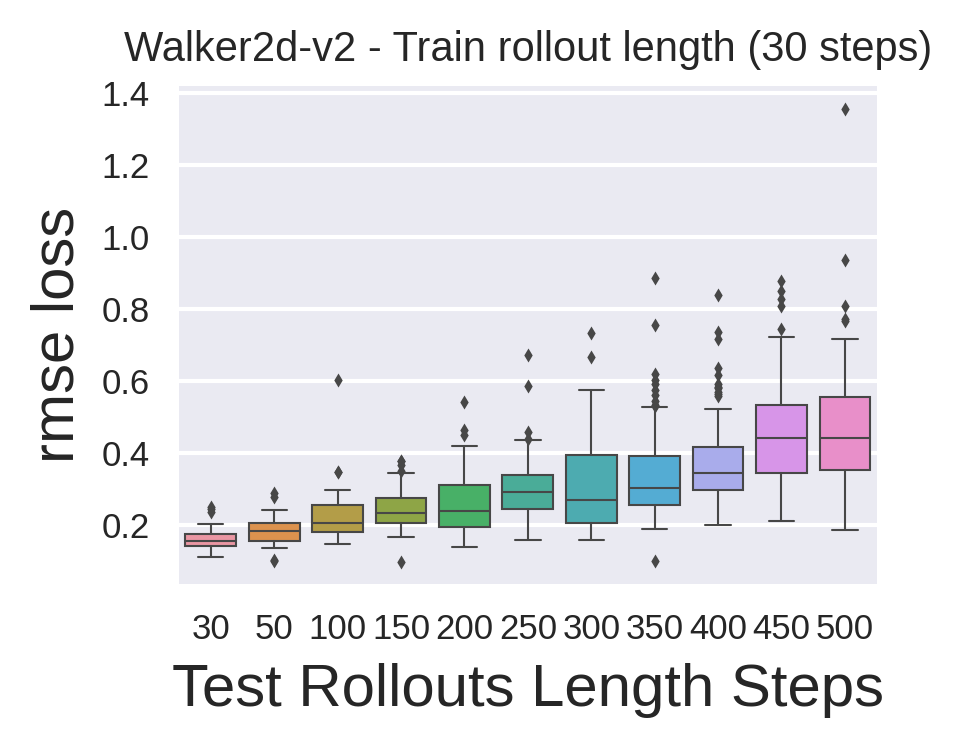}
\includegraphics[width=0.32\linewidth,height=80pt]{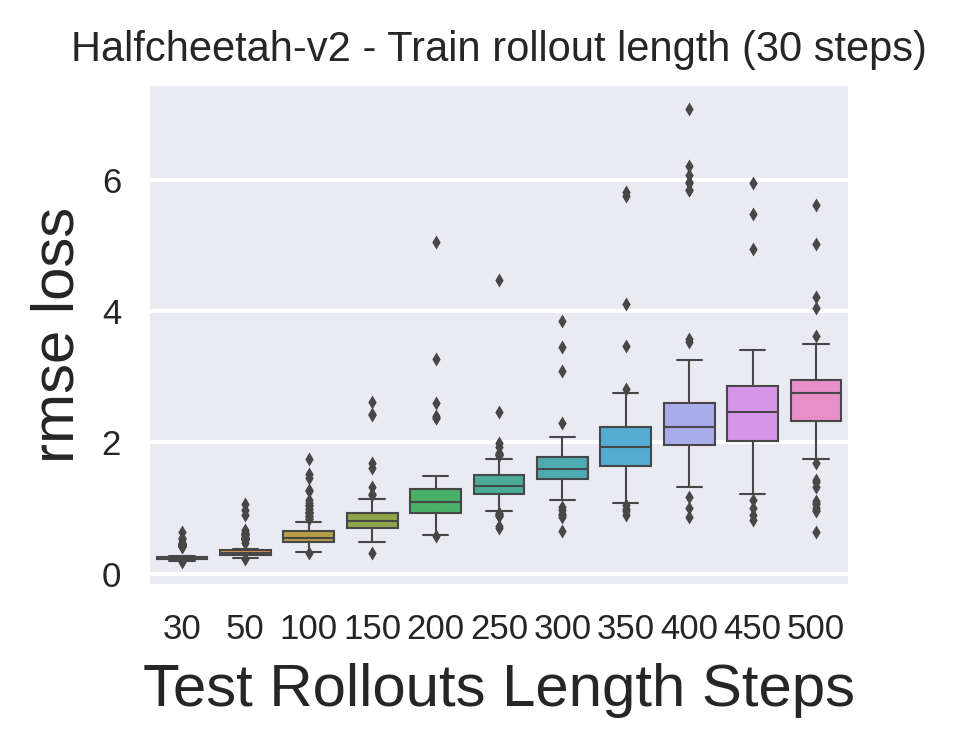}
 
\caption{Hopper, Walker2D and HalfCheetah: RSME per episode and rollout test performance}
\label{App:training_results}
\end{figure}

For Hopper, Figure~\ref{App:training_results} shows the overall RMSE of predicted variables and the total reward for the forward environment. After running a total of 12K episodes and $\sim$121k gradient steps, the overall RMSE converged close to 0.10. Additionally, both actor and critic loss curves show that the SAC agent is learning a good policy from episodes 1000 to 4000 (see \ref{App:training_results} for additional experimental results). After these episodes, both curves converge asymptotically, suggesting a pseudo optimal policy has been achieved. Walker2d has similar number of gradient steps and episodes as Hopper, it is noticeable how errors are higher in terms of RMSE (Fig. \ref{App:training_results}) as well as actor and critic losses (see Appendix \ref{App:training_results}) than Hopper. This is basically due to the higher number of predicted variables (almost double), and the higher complexity of the environment. We can also notice an error increase when the model is tested with higher rollout-steps. However, the asymptotic behavior of these curves is still maintained, suggesting the finding of a stable sub-optimal policy. Finally, HalfCheetah is the hardest environment tested. Despite having the same number of target variables than Walker2d, these are noisier as the dynamic of the robot is more complex and the results seems no to be as good as expected. 

\subsection{Comparison of Reinforcement Learning vs. "vanilla" Supervised Learning}\label{SL_RL_COMPAR}
To compare both Supervised Learning and Reinforcement Learning approaches we selected the Hopper environment and designed a specific setup. 

\paragraph{Reinforcement Learning setup:} Two SAC based RL forward models have been trained (RL-v1 and RL-v2) with different stack windows. RL-v1 uses a stack of $w=10$, while RL-v2 uses $w=30$. The reward function used for this experiments is a fully-sparse reward, that is, $r=0$ for every time step, and $r=$ signal similarity measure (between both true and predicted trajectories) at every rollout end. This has been designed in this way to test the RL approach in it's extreme version, despite this sparse reward function will hurt convergence speed. The signal similarity measure (RL goal) used in this experiments is a simplified but effective similarity measure. It incorporates explicit measures from time-domain, frequency-domain and power-domain measures~\citep[c.f.][]{ivan.similarity2022}. The rollout length for training was $h=50$ episode steps.

\paragraph{Supervised Learning setup:} Two Supervised Learning models (SL-v1 and SL-v2) have been trained and tested with different stack windows as well. SL-v1 has a stack of $w=10$, while SL-v2 uses $w=30$, same as RL setup. The Supervised Learning experiment follows the same structure of steps/rounds of the RL training setup. In particular, we use the same SAC Actor Networks for the Supervised Learning policy, however, samples are taken from the true dataset instead of a replay buffer as RL does. Each sample is of batch-size~$=2^{10}$ and, at every step/round, $100$ mini-batches are randomly sampled to train the network using the Pytorch's MSE Loss.

Thus, both approaches use the same Policy Networks, but trained using two different methods. Special attention have to be paid to the fact that these two approaches solves different problems, since they have different optimization objectives: the RL training uses as its goal a signal similarity function over entire rollouts, while the SL training optimizes the MSE over random mini-batches.

\subsubsection{Training metrics}
\begin{figure}[htb]
 \centering
  \includegraphics[width=\linewidth]{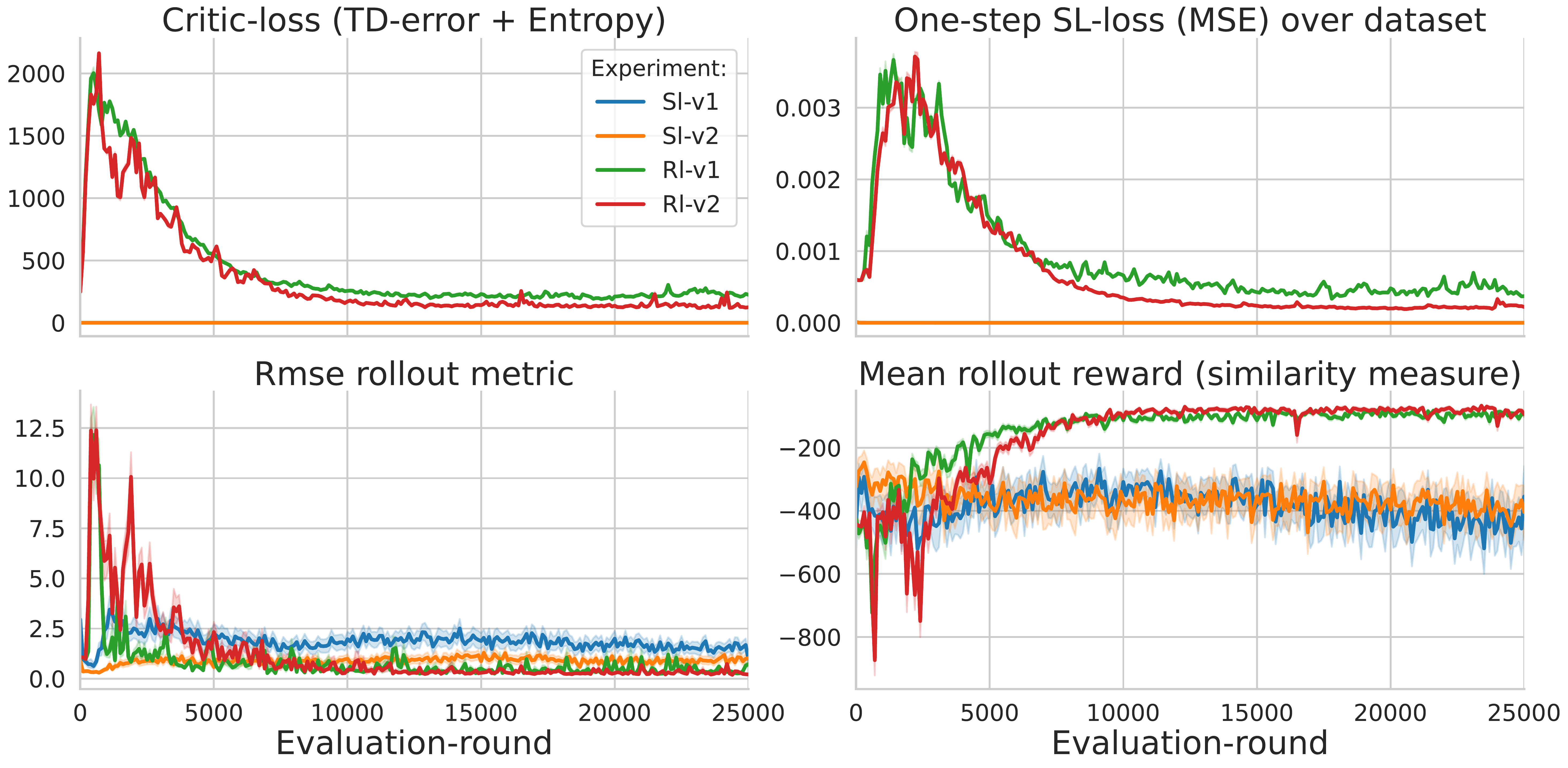}
   \includegraphics[width=0.55\linewidth]{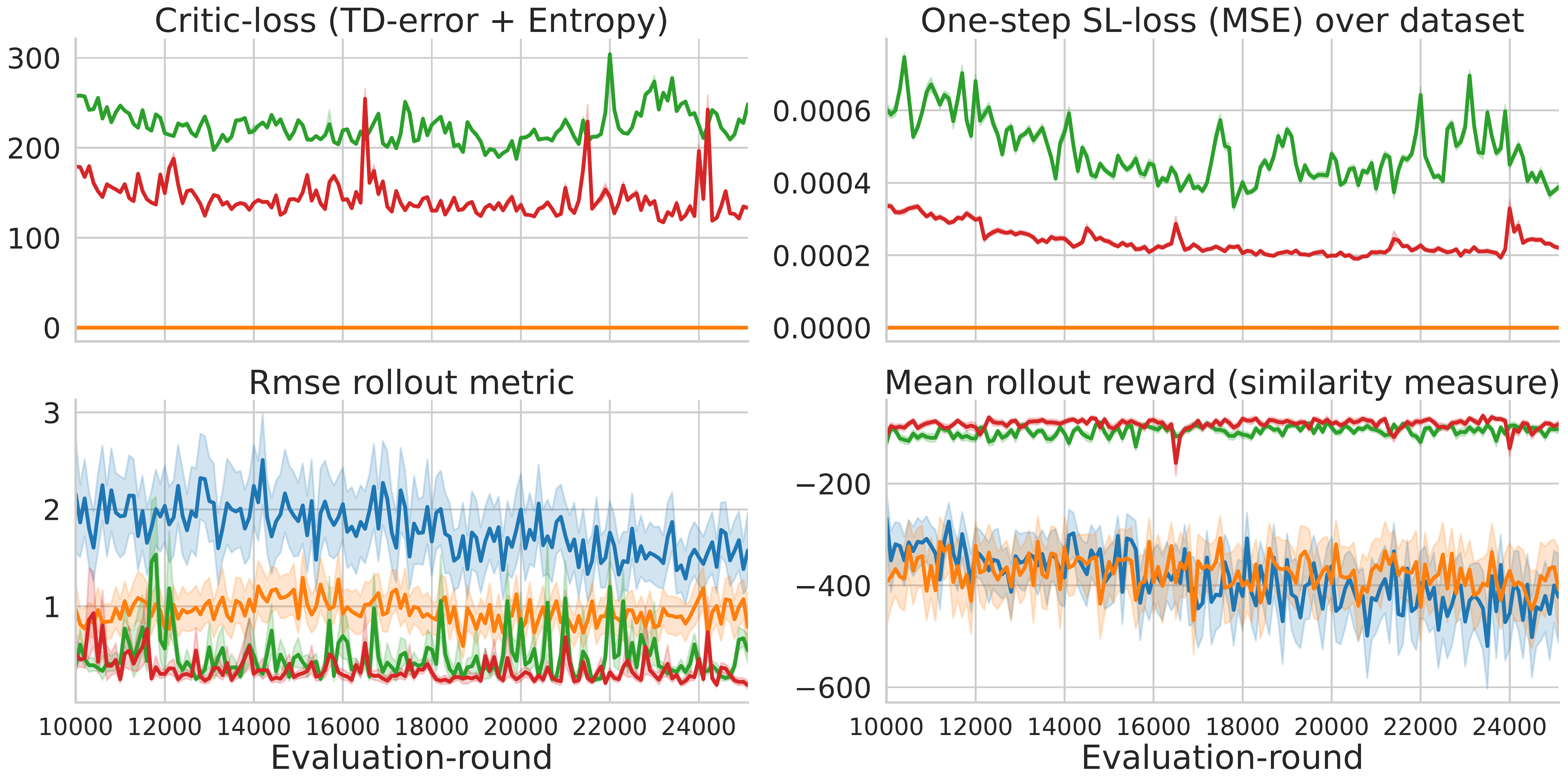}
   \raisebox{9pt}{\includegraphics[width=0.44\linewidth]{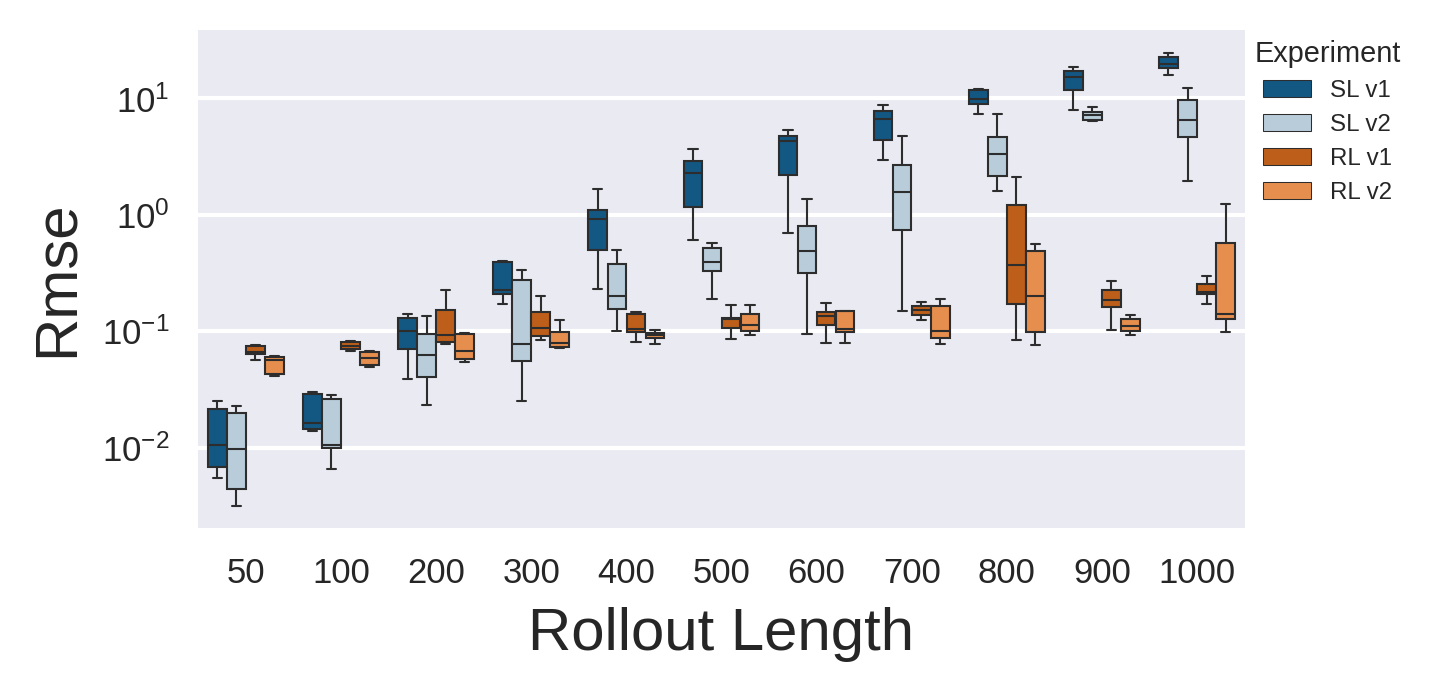}}
\caption{SL and RL Training metrics (upper) and zoom-in last episodes (lower-left). Rollout tests (lower-right) performed with different lengths (logarithmic scale).}
\label{fig:sl_training}
\end{figure}

A total of $25000$ steps/rounds have been performed per experiment. Figure~\ref{fig:sl_training} shows four different plots:
\begin{itemize}
    \item[$\ulcorner$] Critic Loss (upper-left) of the RL approach per evaluation-round. The Critic Loss for SL is $0$ since no critic is trained for the SL experiments.
    \item[$\urcorner$] Supervised loss MSE (upper-right) per evaluation-round for all policies (RL+SL) since it is possible to evaluate an RL-actor policy network over the true dataset in the same way as the SL policy. 
    \item[$\llcorner$] RMSE rollout-metric (lower-left) which is the RMSE evaluated at the current evaluation-round over all the steps of a random episode $e_r$ from the dataset, comparing the predicted trajectory of the entire episode vs. the true trajectory of the entire episode.
    \item[$\lrcorner$] Mean rollout reward (lower-right) shows the mean of the rewards obtained by the evaluation of the signal similarity measure at each rollout end of the random episode $e_r$.
\end{itemize}

Some conclusions can be elaborated from observing the results in Figure~\ref{fig:sl_training}:
\begin{enumerate}
\item SL trained policies converge very fast to very low values of the Supervised loss MSE and low values of the RMSE rollout-metric (SL-v1 and SL-v2 lines almost converge in the firsts training rounds), however, it is not the same for the Mean rollout reward (signal similarity measure) and the SL-v2 version ($w=30$) achieves better optimality than SL-v1 ($w=10$).
\item  The RL trained policies convergence speed is notably slower (orders of magnitude), however both versions of RL policies achieves better optimality than SL in Rmse rollout-metric and Mean rollout reward. Although RL-v2 policy of ($w=30$) shows a better optimality in the RMSE rollout-metric than the RL-v1 ($w=20$), it is not clear that this is true for Mean rollout reward metric.
\item It is somehow natural that the trained SL policies does not perform well for the Mean rollout reward metric, since they are not trained with this learning objective, however, both RL policies are also converging to low enough values of the Supervised loss MSE metric, even when they are not trained with this objective.
\item  Finally, from the whole figure, it can be observed that beyond some "rounds" RL starts to gain advantage over the vanilla SL approach, not only achieving better values in the RMSE rollout-metric and Mean rollout reward metrics, but also with significantly less standard deviations, as shown in the shaded regions enclosing all curves.
\end{enumerate}

\subsubsection{Rollout size evaluation}
With the corresponding trained policies, we tested each one by predicting trajectories on random episodes for different rollout lengths: $h \in \{50,100,200,300,\ldots,1000\}$. According to the results in Figure~\ref{fig:sl_training}, it is noticeable how the SL policies (v1 and v2) perform better than RL for shorter rollouts ($h \lesssim 300$), however, as we increase the length of the rollouts, the SL trained policies  suffer an exponentially increasing compounding error. Nevertheless, the RL approach seems to be more robust against rollout lengths, suggesting, perhaps, that RL polices are able to better grasp the underlying dynamics, which is crucial for predicting full trajectories, especially those that are highly multimodal instead of just the trend following behavior.

\begin{figure}[htb]
\centering
\includegraphics[width=\linewidth]{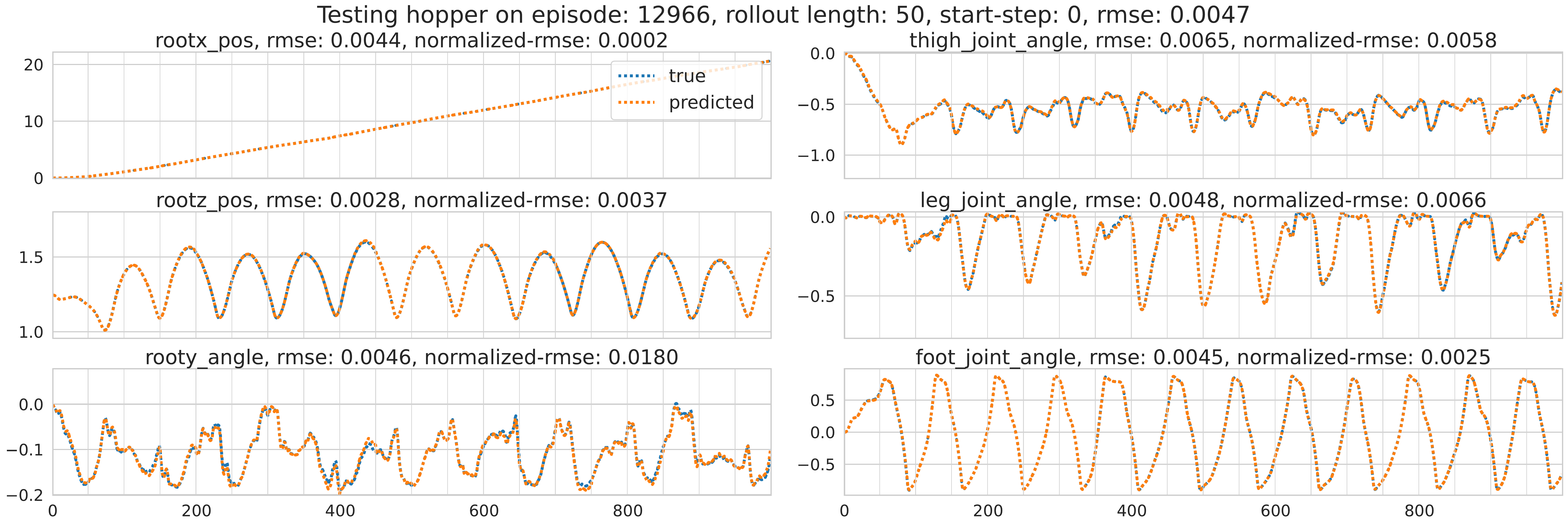}

\hrule

\includegraphics[width=\linewidth]{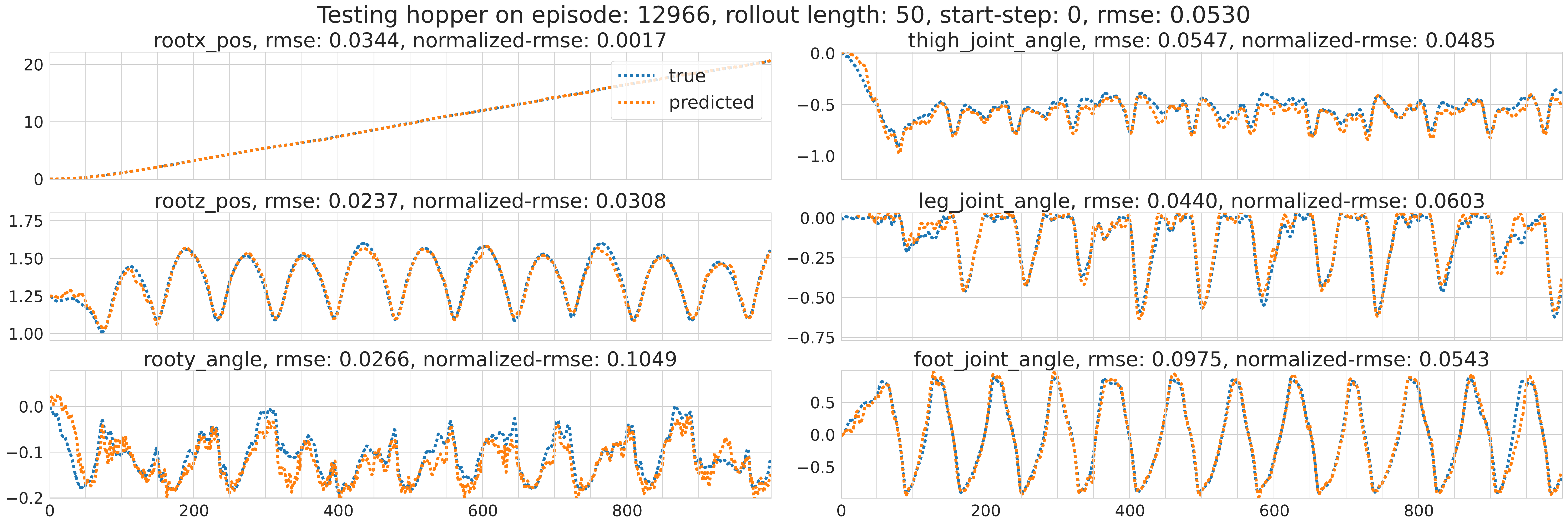}
\caption{Random episodes of rollouts length 50 steps for the Supervised Learning (upper) and Reinforcement Learning (lower) policies.}
\label{fig:rl_sl_episode_renders_50}
\end{figure}

\begin{figure}[htb]
\centering
\includegraphics[width=\linewidth]{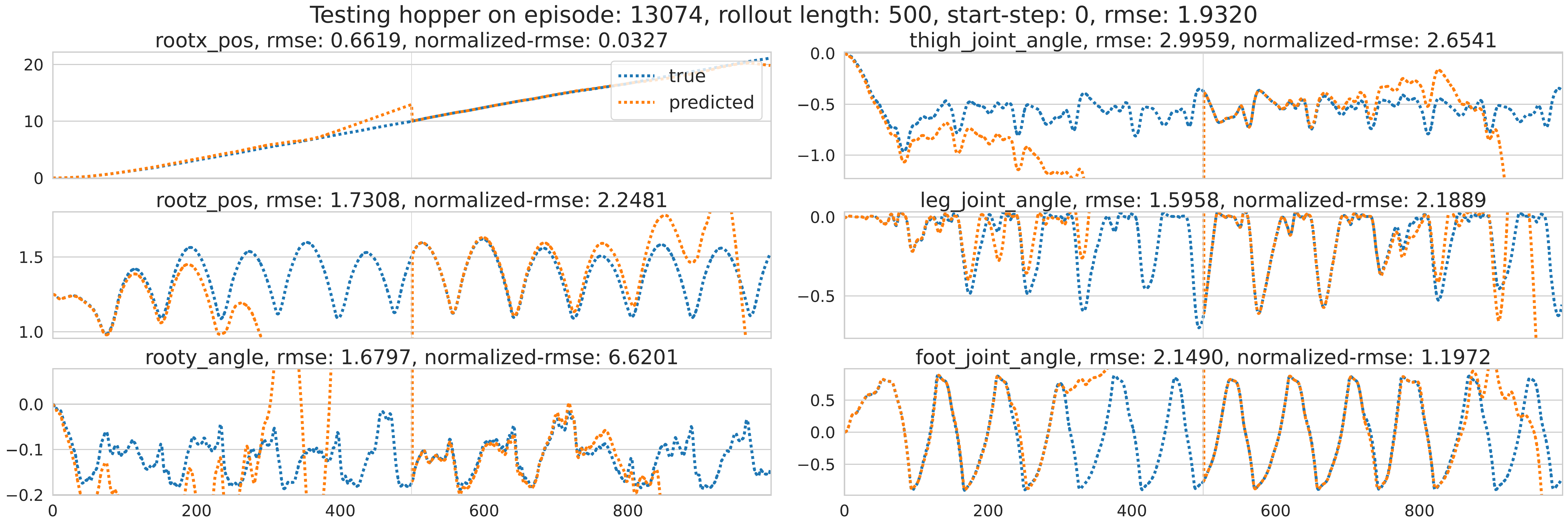}

\hrule

\includegraphics[width=\linewidth]{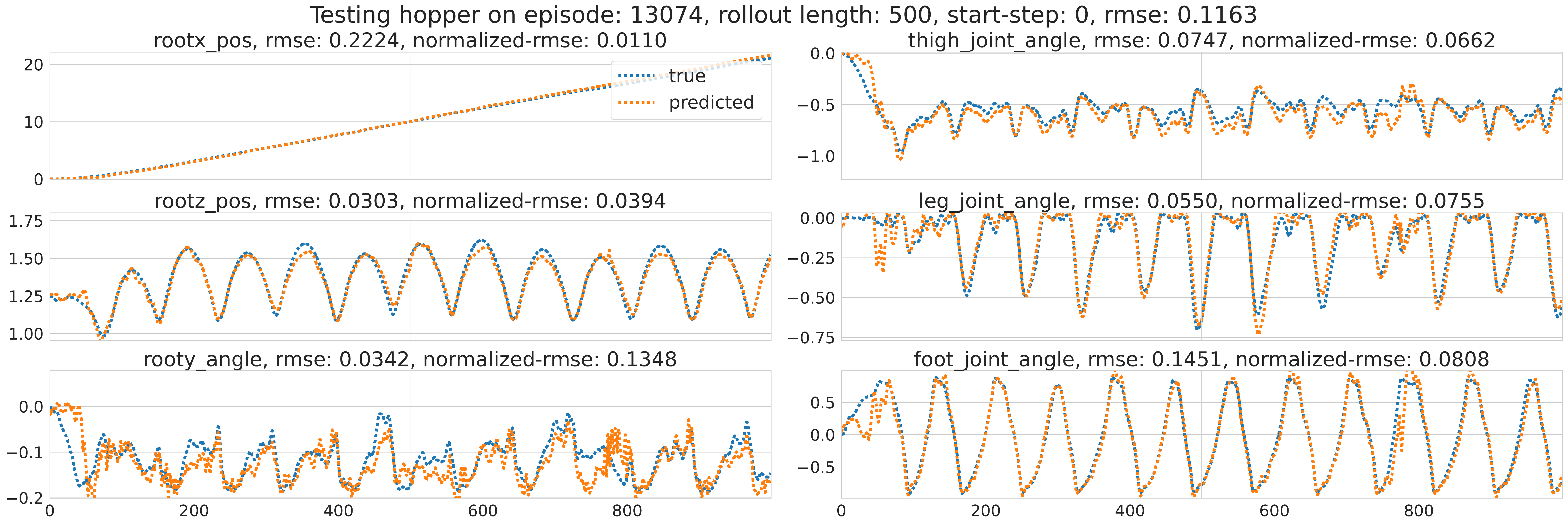}
\caption{Random episodes of rollouts length 500 steps for the Supervised Learning (upper) and Reinforcement Learning (lower) policies.}
\label{fig:rl_sl_episode_renders_500}
\end{figure}

Figures~\ref{fig:rl_sl_episode_renders_50} and~\ref{fig:rl_sl_episode_renders_500} show examples of Hopper's random episodes showing the true vs. predicted trajectories using rollouts of $50$ and $500$ steps, for both SL-v2 and RL-v2 policies.

\section{Conclusion and future work} \label{sect:Conclusions}
This paper presents a general framework to learn forward models with Reinforcement Learning. We have shown significant reasons to consider forward-model learning as an RL problem instead of the common definition as a Supervised Learning one. We have also defined and described how to model this problem with RL, and finally, the proposed method was tested over three well-known environments from the MuJoCo Gym collection, and compared against a commonly-used Supervised Learning framework, showing significant results in reducing the compounding error for large horizon simulations. We argue that main reasons behind are: 1. RL controls the compounding error, minimizing it in the long-run by solving the temporal credit assignment problem; 2. That using a similarity function on whole rollout trajectories is a preferred objective over the widely adopted $\ell_p$-norms; 3. RL exploration during training on bootstrapped rollouts enriches the robustness of the policy to noise and the compounding errors as well. 

The guiding objective is the use of these forward models to train RL control agents. This will allow exploration in offline RL, which is a very desirable feature to get optimal and robust enough policies, and thus compare more fairly their performance against agents trained in "real" environments. 

A collateral contribution to consider is that forward-model learning can be used as new kind of benchmark for RL algorithms, enforcing the requirements for fast but deep exploration, convergence-optimization and generalization. Using RL for forward-model learning also brings homogeneity to model-based RL by, perhaps, inspiring new architectures with mixed $Q$-functions and policies.



\clearpage
\bibliography{neurips_2022}
\clearpage
\newpage
\appendix

\section{Appendix}

\subsection{Extended discussion and concluding remarks}

\paragraph{Real Life motivation} Reinforcement Learning is a very powerful technique for solving Industrial Control Problems. However, in the Industrial setting, there are many mission critical assets where RL cannot be applied in its canonical form, for security or operational risks, i.e., the risk of unsafe exploratory actions or the interruption of asset operation. Among the last lines of research on this matter, the "Offline Reinforcement Learning" formulation drops many barriers in the successful application of RL to these problems. Offline RL uses observations from the true industrial asset operation trough a dataset of trajectories (including control actions). This dataset is then used to perform offline RL to learn a control policy.

However, it is very common, by causes of operational reasons, that not all control actions are well represented in the dataset (out of distribution actions --OOD actions--), and thus the estimated value of such actions may be largely over/under estimated. The reason is that the returns of such OOD actions can not be learnt (penalized/rewarded) during training, since there is no interaction with the true environment. Hence, OOD actions become untested actions with maximal uncertainty in its estimated values. This fact, induces poor (or completely wrong) policies withing the canonical RL setting, and the performance of the current state of the art Offline RL algorithms is being actively investigated. However, if a forward model is provided then RL algorithms can explore on it, and avoid or mitigate the problem of ODD actions.

Along this line, the natural improvement over the presented results, is being able to understand how good or bad our forward models represent the real environment dynamics (beyond the information contained in source datasets and error metrics). In order to analyse this key point, the goal is to train standard RL agents on these models and compare their performance against real environment trained ones.

Although we have done initial steps on this direction, setting up, and controlling the experimental setup of such experiments must be further studied, since our initial tests did not yield the expected results. We argue that once these tests are completed, their results will give us a better understanding of the robustness of our approach.

\paragraph{The claim} First, we want to remark that the presented comparison between Supervised Learning and Reinforcement Learning is not about algorithms, but problem paradigms. We do not deny that Supervised Learning, with enough extra features, can be used to solve the problem as effectively as any Reinforcement Learning approach or even better. However, what we argue is that the procedure to build such specialized Supervised Learning method signals the quest to solve a problem that is more naturally stated as a Reinforcement Learning problem or a general sequential decision problem.

\paragraph{The proposed method} There are many other aspects to consider for further optimization of the learning procedure, for instance:

\paragraph{Does this problem impose a special constraint to the $Q$-function?} Since we know that a sequence of perfect actions should return $0$ as rollout error, it can be argued that we know in advance the right returns of many $(s,a)$ pairs if we do not consider signal noise and a source of randomness in the collected data. We can even consider that these sources exist, but are negligible so we can then explicitly impose such constraints into the learning of the $Q$-function.

Along this line, one tested trick is to select random tuples $(s, a, s', r=0)$ from the datasets and update the critic network with such tuples with the aim of improving learning convergence. However, in our initial experiments, using this trick cause the $Q$-function to diverge if the ratio of this updates vs. the updates from the replay buffer is not controlled.  

\paragraph{Transferring the knowledge from the policy to the $Q$-function} Another trick is to try training the policy network via Supervised Learning with some random $(sa,s')$ pairs so that the policy can take advantage of Supervised Learning as well. But this will require a special procedure to update the critic network specifying confidence on some $(s,a,R)$ tuples. This looks analogous to the procedures used by offline-RL algorithms to constraint the $Q$-values on out-of-distribution (OOD) points. We have tested this trick by helping the policy learning using random $(sa,s')$ to try to improve learning, however without a clear procedure to update the critic network as well, there is a fierce competition between the updates to the policy to optimize the critic and the Supervised Learning ones. Besides, these results show also a divergence in the $Q$-function and the policy learning.

\paragraph{Avoiding the hacker agent against a learned model} A known common concern is the idea that an agent trained over a learned forward model may exploit weaknesses of the model to gain unfair advantage in optimizing the reward function (the cheating effect). Some authors have argued that this can be alleviated due to the inherent noise contained in the learned forward-model. Another simple idea is to over constraint the reward function for the learned forward models. We think that a future research direction is how to prevent this cheating effect. However, in our early experiments we can not conclude that the agents systematically gain unfair advantages. Instead, what we observe is just slightly different behaviors that of course yield different returns, but not a clear trend to optimize over the model, degrading the performance in the original environment.

\clearpage

\subsection{Hardware and Software} \label{App:hardware}
For all the experiments we have used a Linux virtual machine with enough system RAM and $4$ Nvidia Tesla-T4 GPUs. The experiments presented here rely on the following Deep Reinforcement Learning specific software: for the environments (problems to solve) OpenAI Gym~(\citet{openaigym}) and MuJoCo~(\citet{mujoco}) are used. For Deep RL algorithms, the SAC implementation in the d3rlpy library~(\citet{seno2021d3rlpy}) is used. For implementing a replay buffer we do not use the d3rlpy's replay buffer implementation, instead the cpprb library~(\citet{Yamada_cpprb_2019}) is used. Finally, as the Deep Learning backend and framework, Pytorch~(\citet{pytorch}) is used. D4RL datasets (\cite{fu2020d4rl}) are used as as the source of dynamics trajectories. Every library and asset mentioned in this paper and used in this research has the license to be used without restriction.

\subsection{Environments' spaces} \label{App:EnvSpaces}

In this section, we present the list of variables of the state spaces for both the original problem $\mathcal{M}$ and the forward model approach $\mathcal{M_F}$ for the three MuJoCo environment used in the experimental evaluation. We recall that in the case of the forward model environment we have decided to predict only the position's state variables (target variables) because the velocities can be derived from them in order to reduce complexity of the problem. Those variables have been marked with an (*).

\begin{table}[htb]
  \caption{State space variables for MuJoCo and derived forward environments }
  \label{table:mujoco_variables}
  \centering
  \begin{tabular}{lll}
    \toprule
    Hopper     & Walker2d     & Halfcheetah  \\
    \midrule
rootx pos*             	&	rootx pos*	                &	rootx pos*	\\
rootz pos*	            &	rootz torso*                &	rootz pos*	\\
rooty angle*            &	rooty torso angle*	        &	rooty pos*	\\
thigh joint angle*	    &	thigh joint angle*	        &	bthigh angle*	\\
leg joint angle*   	    &	leg joint angle*            &	bshin angle*	\\
foot joint angle*	    &	foot joint angle*	        &	bfoot angle*	\\
                	    &	thigh left joint angle*	    & 	fthigh angle*	\\
	                    &	leg left joint angle*	    &	fshin angle*	\\
                        &	foot left joint angle*	    &	ffoot angle*	\\
\hline
rootx vel	           &	rootx vel	                &	rootx vel	\\
rootz vel              &	rootz vel	                &	rootz vel	\\
rooty angle vel        &	rooty angle vel	            &	rooty angle vel	\\
thigh joint angle vel  &	thigh joint angle vel	    &	bthigh angle vel	\\
leg joint angle vel	   &	leg joint angle vel	        &	bshin angle vel	\\
foot joint angle vel   &	foot joint angle vel	    &	bfoot angle vel	\\
	                    &	thigh left joint angle vel	&	fthigh angle vel	\\
	                    &	leg left joint angle vel	&	fshin angle vel	\\
	                    &	foot left joint angle vel	&	ffoot angle vel	\\
    \bottomrule
  \end{tabular}
\end{table}

\newpage
\clearpage

\subsection{Training data preparation} \label{App:datasets}

Training data for the forward model have been extracted from D4RL repositories (\cite{fu2020d4rl}). For each example, several datasets from this library have been included in order to get a large and diverse collection of trajectories (see Table \ref{table:files-table}). Additionally, and for training purposes they have been filtered only those experiences with episodes longer than a minimum number of steps (500 for the main experimental results and 100 for the additional tests). This filtering aims to find longer and more stable episodes to learn.

\begin{table}[htb]
  \caption{D4RL source datasets used per experiment.}
  \label{table:files-table}
  \centering
  \begin{tabular}{lll}
    \toprule
    Hopper-v2     & Walker2d-v2     & Halfcheetah-v2  \\
    \midrule
hopper-random-v2 &	walker2d-random-v2 &	halfcheetah-random-v2 \\
hopper-medium-v2 &	walker2d-medium-v2 &	halfcheetah-medium-v2 \\
hopper-expert-v2 &	walker2d-expert-v2 &	halfcheetah-expert-v2 \\
hopper-medium-replay-v2	& walker2d-medium-replay-v2 &	halfcheetah-medium-replay-v2  \\
hopper-medium-expert-v2	& walker2d-medium-expert-v2 &	halfcheetah-medium-expert-v2 \\

    \bottomrule
  \end{tabular}
\end{table}

\subsection{Training procedure}
\label{App:training}
Here, you can find the main algorithms of the presented forward models (Algorithms \ref{alg:fw-model}, \ref{alg:fw-model-collect}, and \ref{alg:fw-model-env}).

\begin{algorithm}
   \caption{Forward model main training loop}
   \label{alg:fw-model}
\begin{algorithmic}[1]
    \STATE Initialize replay-buffer ($RB$), SAC algorithm ($SAC$) and forward model GYM  environment ($FW$)
    \STATE
    \STATE total steps = 0
    \FOR{$episode$}
        \STATE steps = 0
        \STATE total reward = 0
        \STATE results = collect($FW$, $SAC$) \COMMENT{(described at algo.\ref{alg:fw-model-collect})}   
        \FOR{sample \ in \ results}
            \STATE{$RB$ $\gets$ sample } \COMMENT{(append)}
            \STATE{steps $+=$ sample[steps]  }
            \STATE{total reward $+=$ sample[reward]  }
        \ENDFOR
        \STATE {total steps $+=$ steps}
        \FOR{$i$ in range(10)}
             \STATE{samples = $RB$.sample($SAC$.batchsize)  }
             \STATE{loss = $SAC$.update(samples)}
        \ENDFOR
        \STATE{Update \ metrics}
        \IF{$episode$  $\%$ $100$ == $0$}
            \STATE Save SAC model and policy
        \ENDIF
   \ENDFOR
\end{algorithmic}
\end{algorithm}

\begin{algorithm}
   \caption{Collect Algorithm}
   \label{alg:fw-model-collect}
\begin{algorithmic}[1]
    \STATE Initializate $buffer$. env and SAC included as argument.
    \STATE
    \STATE state = env.reset()
    \FOR{$n$ in $range(10000)$}
    \IF{explore}
        \STATE action = SAC.sample(state)
    \ELSE{}
        \STATE action = SAC.predict(state)
    \ENDIF
    \STATE {step = env.step(a)}  \COMMENT{described at algo.\ref{alg:fw-model-env}}
    \STATE state = step[s]
    \STATE {$buffer \leftarrow step$} \COMMENT{append}

   \ENDFOR
   \RETURN{ $buffer$}
\end{algorithmic}
\end{algorithm}

\begin{algorithm}
   \caption{Forward model environment step algorithm}
   \label{alg:fw-model-env}
\begin{algorithmic}[1]
  
    \STATE {Initializate observation as $stack$, predicted state as $buffer$ and true state as $buffer$. dataset included as argument, with the real experiences from d4rl datasets. action performed by SAC agent (sac action) is passed as argument.}
    \STATE
    \STATE step counter $+=$ 1
    \STATE rollout step counter $+=$ 1
	\STATE true obs = dataset[step counter][observation]
	\STATE true action = dataset[step counter][action]
	\STATE predicted state = observation $+$ sac action	
	\STATE observation $=$ predicted state	
	
	\IF{rollout step counter $>=$ rollout step}
		\STATE observation $=$ true obs 
		\STATE rollout step counter $=$ 0
		\STATE rollout terminal $=$ True
	\ENDIF
	\STATE predicted $\leftarrow$ predicted state, action, rollout terminal \COMMENT{append}
	\STATE true state $\leftarrow$ true obs, true action, terminal	 \COMMENT{append}
	\IF{step counter $>=$ total episode length}
		\STATE terminal = True
	\ELSE
		\STATE terminal = False
	\ENDIF
	\STATE{reward = getReward(predicted state, true state, terminal)}
    \RETURN{observation, reward, terminal}

\end{algorithmic}
\end{algorithm}

\subsection{Additional experimental results} 

In this section, there are some detailed additional results that may result useful in order to understand not only the training stage (i.e., actor and critic losses figures, as well as detailed training main metrics Table \ref{table:param-table}), but also the testing of our forward model on dataset actions. 

\begin{figure}[htb]
\minipage{0.48\textwidth}
  \includegraphics[width=\linewidth]{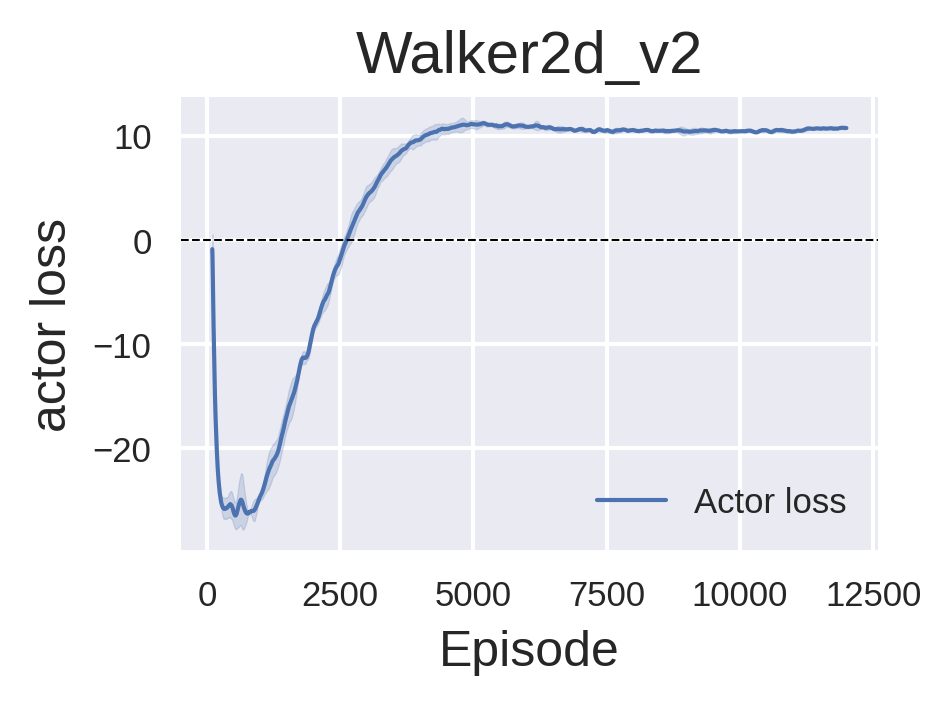}
\endminipage\hfill
\minipage{0.48\textwidth}
  \includegraphics[width=\linewidth]{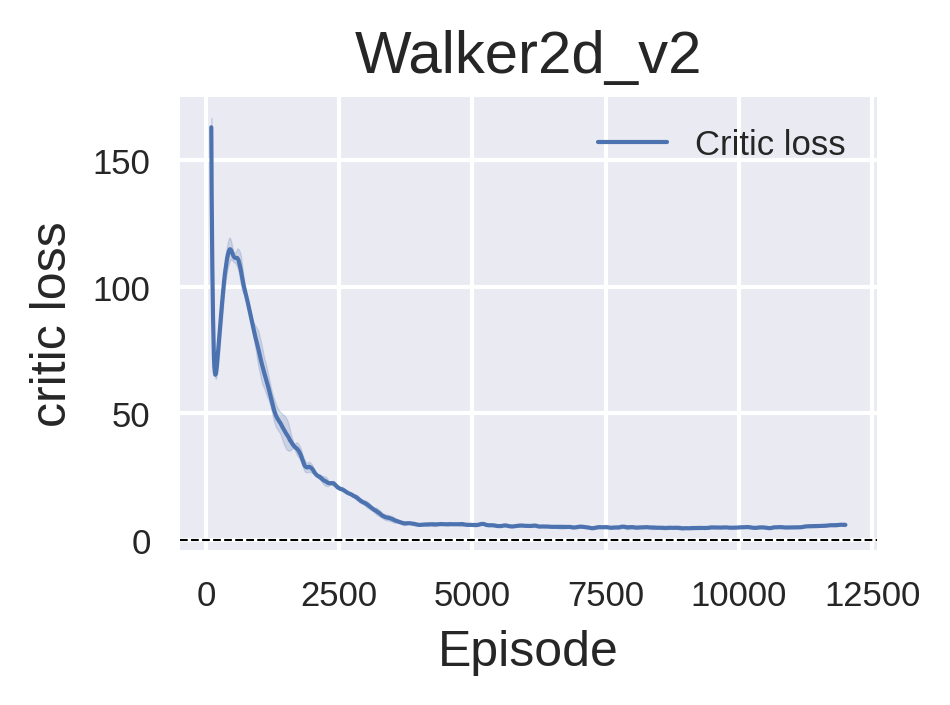}
\endminipage\hfill
\caption{Walker2d SAC Actor loss and Critic loss (training) per episode}
\label{fig:walker_sac_metrics}
\end{figure}

\begin{figure}[htb]
\minipage{0.48\textwidth}
  \includegraphics[width=\linewidth]{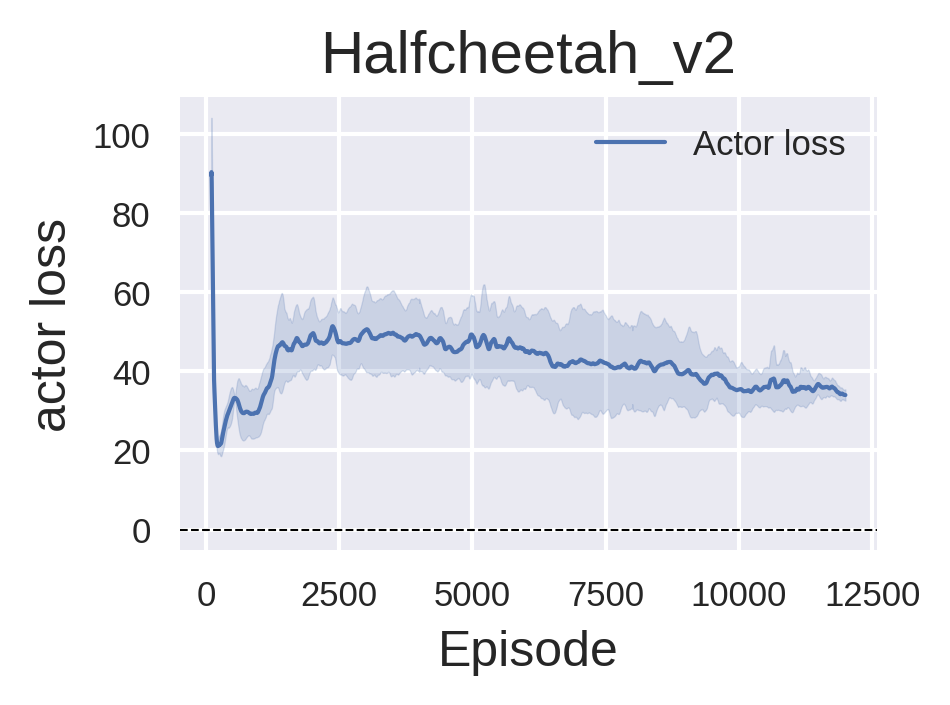}
\endminipage\hfill
\minipage{0.48\textwidth}
  \includegraphics[width=\linewidth]{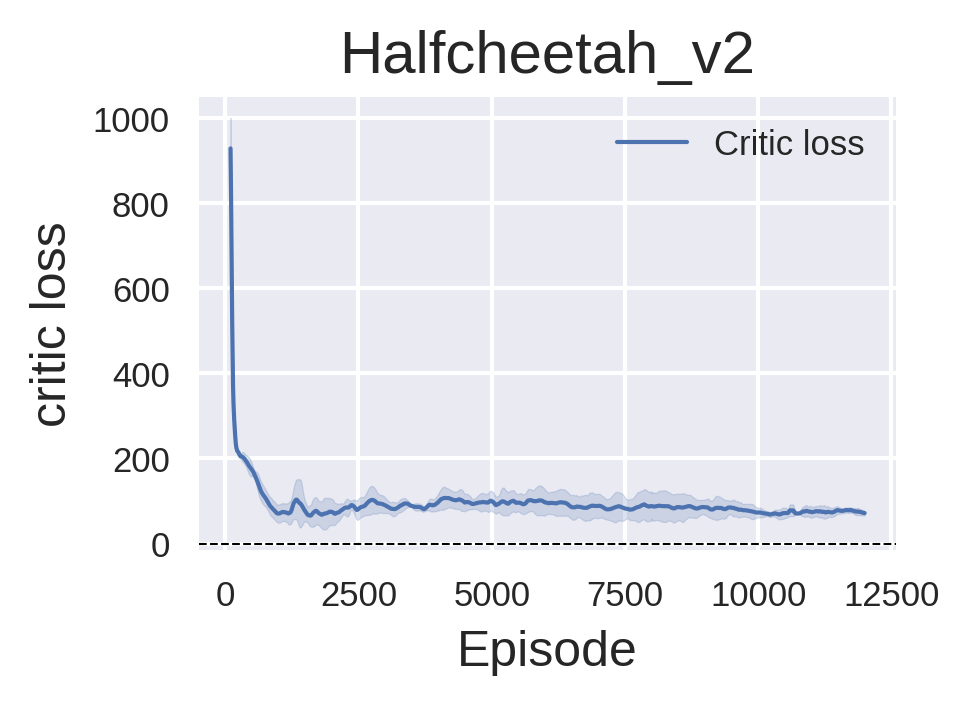}
\endminipage\hfill
\caption{HalfCheetah SAC Actor loss and Critic loss (training) per episode}
\label{fig:cheeta_sac_metrics}
\end{figure}

\begin{figure}[htb]
\includegraphics[width=\linewidth]{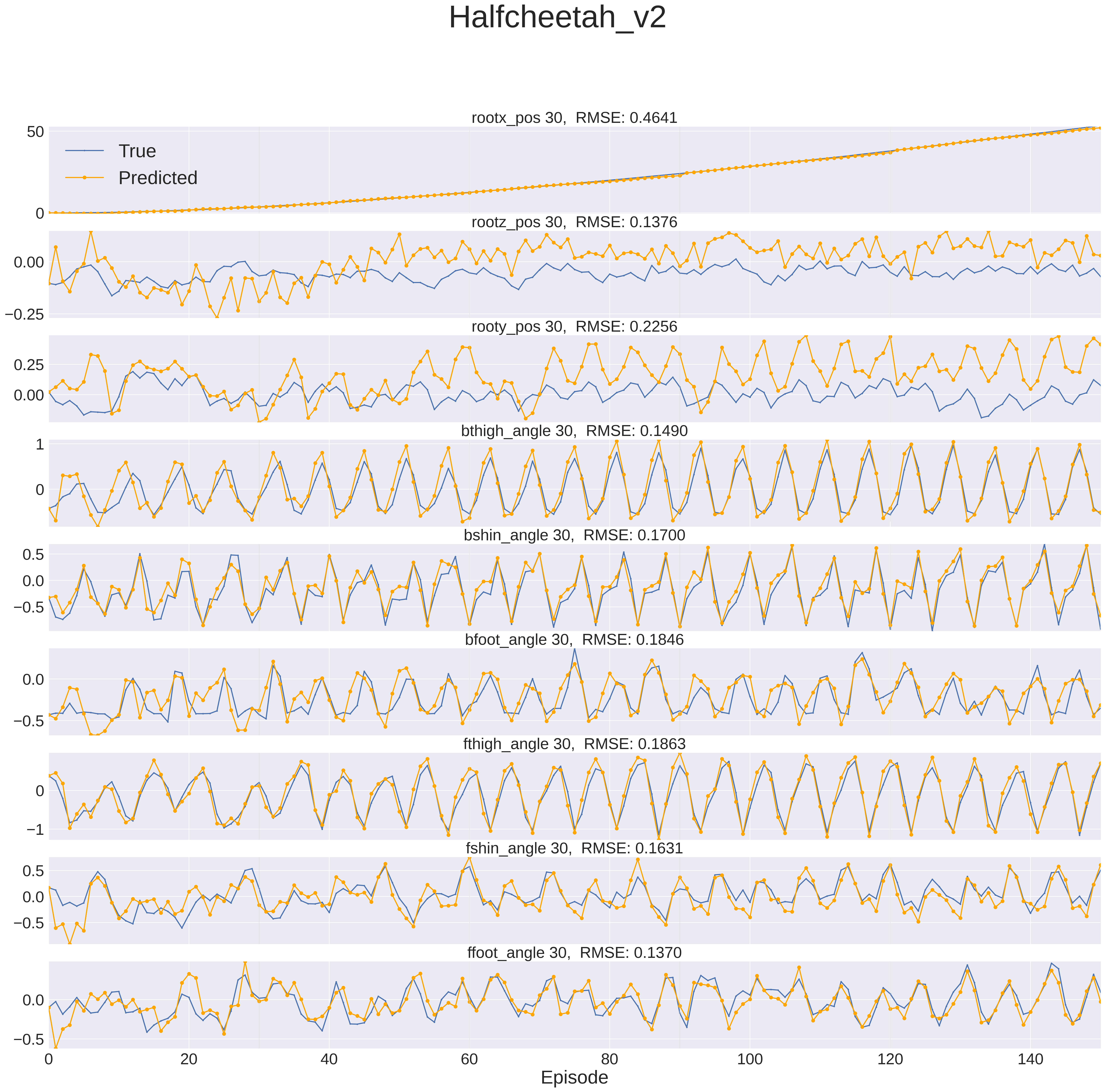}
\caption{Halfcheetah forward model prediction of target variables, compared to real values. Rollouts of 30 steps. Sample of 150 steps (5 full rollouts of 30 steps each) obtained from a full random episode. }
\label{variables-halfcheetah}
\end{figure}

\begin{figure}[htb]
\includegraphics[width=\linewidth]{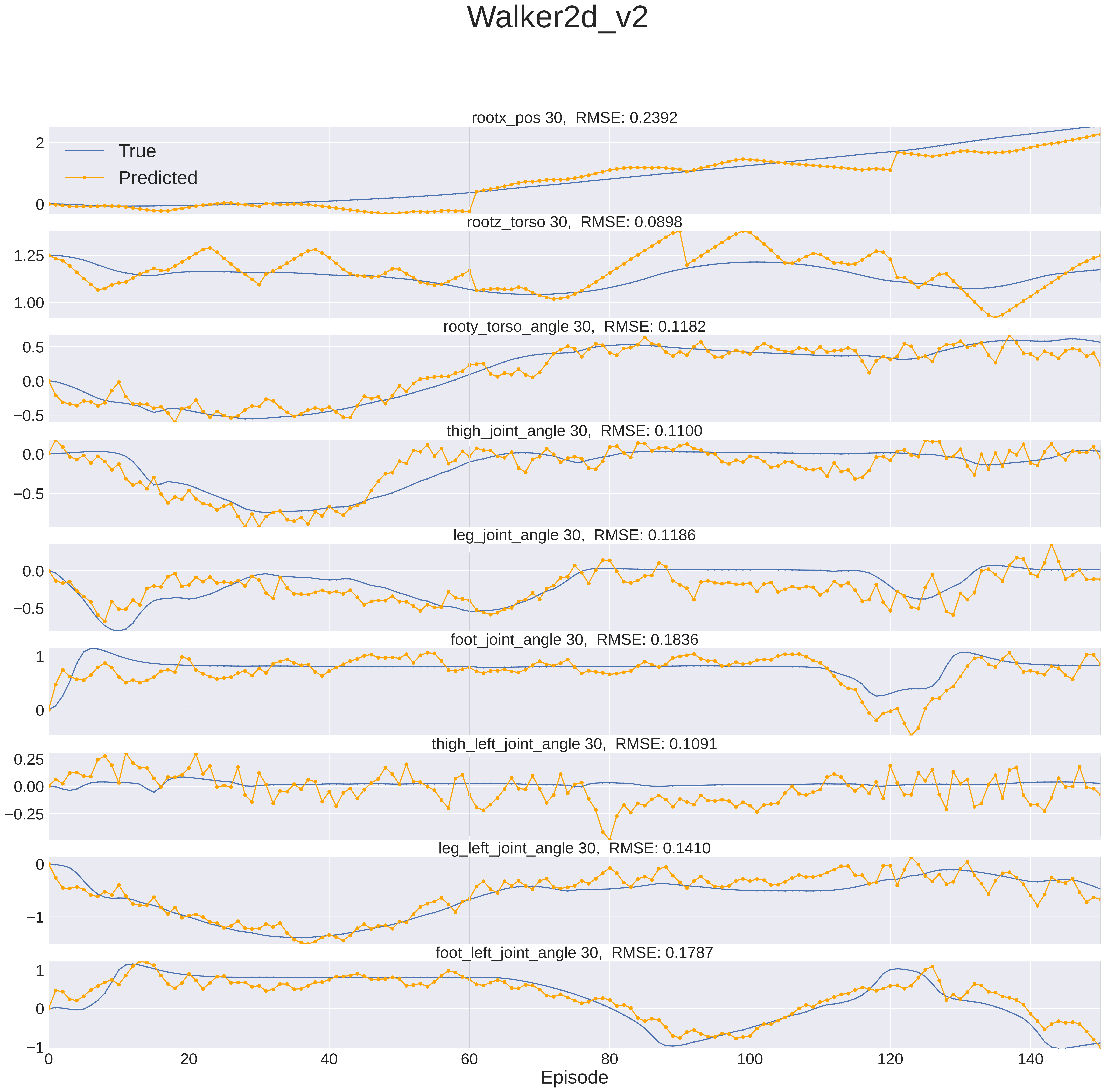}
\caption{Walker3d forward model prediction of target variables, compared to real values. Rollouts of 30 steps. Sample of 150 steps (5 full rollouts of 30 steps each) obtained from a full random episode. }
\label{variables-walker2d}
\end{figure}

\begin{figure}[htb]

\includegraphics[width=\linewidth]{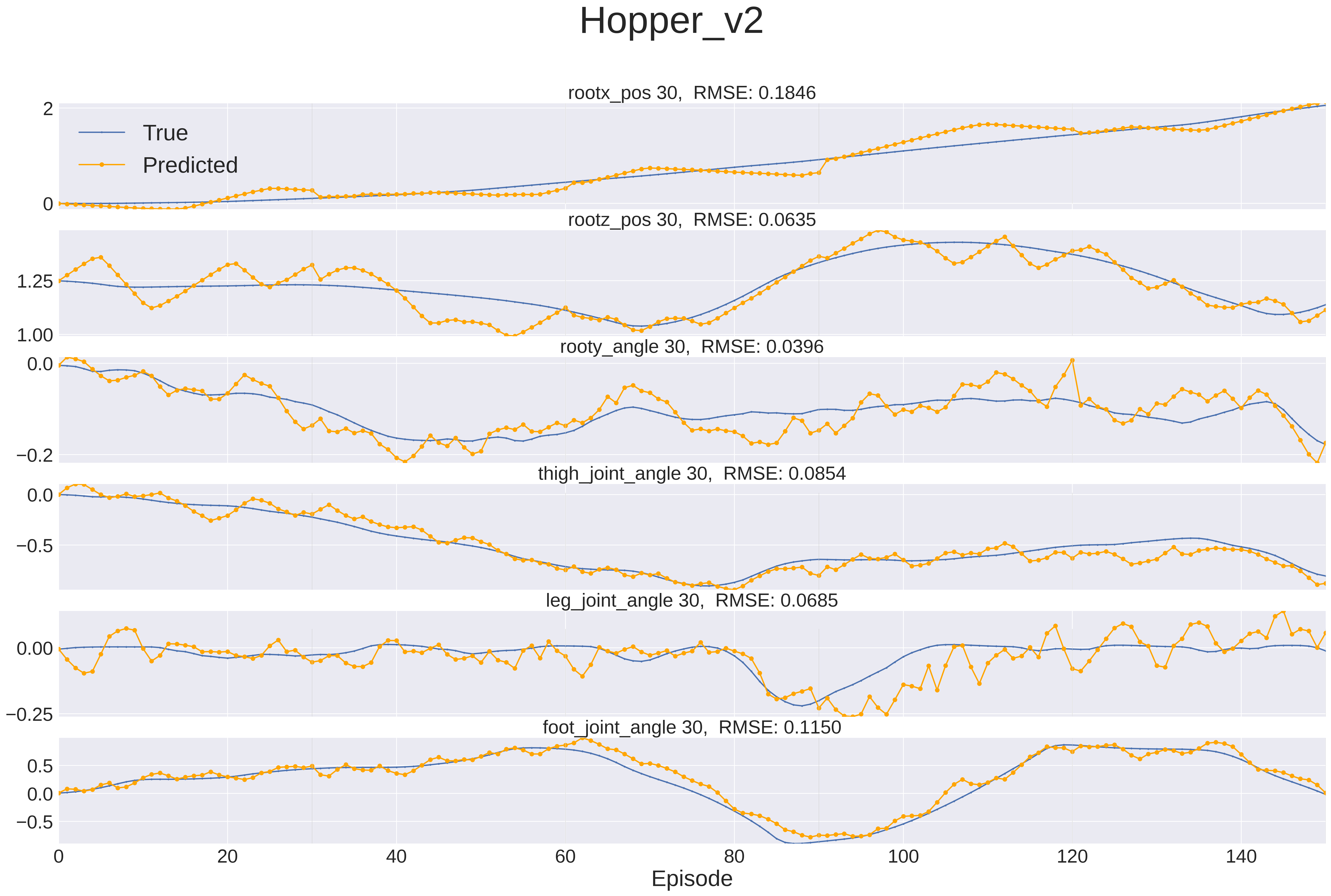}
\caption{Hopper forward model prediction of target variables, compared to real values. Rollouts of 30 steps. Sample of 150 steps (5 full rollouts of 30 steps each) obtained from a full random episode. }
\label{variables-hopper}
\end{figure}

\newpage
\clearpage

\end{document}